# Reconstructing physiological signals from fMRI across the adult lifespan

Shiyu Wang*[a], Ziyuan Xu*[b], Laurent M. Lochard[c], Yamin Li[b], Jiawen Fan[d], Jingyuan E. Chen[d,e], Yuankai Huo[b], Mara Mather[f], Roza G. Bayrak[b], Catie Chang[a,b,c]

[a]Department of Biomedical Engineering, Vanderbilt University, Nashville, TN, USA; [b]Department of Computer Science, Vanderbilt University, Nashville, TN, USA; [c]Department of Electrical and Computer Engineering, Vanderbilt University, Nashville, TN, USA; [d]Athinoula A. Martinos Center for Biomedical Imaging, Massachusetts General Hospital, Boston, MA, USA; [e]Department of Radiology, Harvard Medical School, Boston, MA, USA; [f]Leonard Davis School of Gerontology, Department of Psychology, Department of Biomedical Engineering, University of Southern California, Los Angeles, CA, USA

*These two authors contributed equally to this work.*

## ABSTRACT

Interactions between the brain and body are of fundamental importance for human behavior and health. Functional magnetic resonance imaging (fMRI) captures whole-brain activity noninvasively, and modeling how fMRI signals interact with physiological dynamics of the body can provide new insight into brain function and offer potential biomarkers of disease. However, physiological recordings are not always possible to acquire since they require extra equipment and setup, and even when they are, the recorded physiological signals may contain substantial artifacts. To overcome this limitation, machine learning models have been proposed to directly extract features of respiratory and cardiac activity from resting-state fMRI signals. To date, such work has been carried out only in healthy young adults and in a pediatric population, leaving open questions about the efficacy of these approaches on older adults. Here, we propose a novel framework that leverages Transformer-based architectures for reconstructing two key physiological signals – low-frequency respiratory volume (RV) and heart rate (HR) fluctuations – from fMRI data, and test these models on a dataset of individuals aged 36-89 years old. Our framework outperforms previously proposed approaches (attaining median correlations between predicted and measured signals of r ~ .698 for RV and r ~ .618 for HR), indicating the potential of leveraging attention mechanisms to model fMRI-physiological signal relationships. We also evaluate several model training and fine-tuning strategies, and find that incorporating young-adult data during training improves the performance when predicting physiological signals in the aging cohort. Overall, our approach successfully infers key physiological variables directly from fMRI data from individuals across a wide range of the adult lifespan.

## 1. DESCRIPTION OF PURPOSE

Functional magnetic resonance imaging (fMRI) captures brain activity in an indirect, non-invasive manner by recording changes in the local blood oxygenation level (blood-oxygen level dependent signal; BOLD) [1–3]. Prior work has demonstrated that bodily processes have a close connection with fMRI BOLD signals. For example, slow, natural variations in respiration volume (RV) and heart rate (HR) have been associated with low-frequency fMRI signals (0.01-0.15 Hz [4–6]), a band that is of high relevance for studying brain activity and connectivity[7,8]. Therefore, collecting respiration and heart rate signals during fMRI is important for interpreting fMRI results. Moreover, there is increasing evidence that the coupling between fMRI and physiological signals provides valuable information regarding brain vasculature and cognitive performance[9–13]. However, many fMRI datasets lack simultaneously recorded physiological measures[14–16], which require additional equipment and setup during the fMRI session. Further, even when physiology is monitored, the recordings can be heavily corrupted by artifacts due to hardware, subject movement, or other sources.

To tackle this problem, previous studies have investigated the possibility of reconstructing low-frequency RV and HR

signals directly from fMRI data. These studies have employed convolutional neural networks (CNN) [17,18], bidirectional long short-term memory (Bi-LSTM), and vanilla transformer models [19,20] in healthy young adults (<36 yo) and children (<18 yo). However, to our knowledge, no study has yet modeled aging cohorts (i.e., adults older than 36 yo). There is strong evidence that the relationship between physiological fluctuations and fMRI may change with aging [21–24], suggesting the need for models that operate on older adults. Since its inception, transformer models have achieved excellent performance in capturing short/long temporal dependencies [25]. One candidate approach for reconstructing physiological time-series from fMRI may involve the use of transformer architectures with multi-head sliding-window attention mechanism [26–28]. Here, we propose a novel framework that adapts two transformer-based models to the problem of extracting physiological signals directly from resting-state fMRI data. We find that these models outperform previous CNN and Bi-LSTM based models on an older (>36 yo) adult cohort. In addition, we examine the influence of age on the physiological signal reconstruction performance, and find that including training data from older adults yields no significant age-related performance differences.

The rest of the paper is structured as follows: 1) the *Methods* section provides a brief introduction of the dataset, preprocessing steps, model architectures and training details; 2) the *Results* section presents the RV and HR reconstruction results; and 3) the *Conclusion* section highlights the implications and summary of our work.

## 2. METHODS

### 2.1 Datasets and fMRI preprocessing

We included 1500 resting-state scans with high-quality physiological recordings from 375 subjects (each scanned 4 times) in the Human Connectome Project (HCP) Young Adult dataset ("HCP-Young") and 1752 scans from 571 subjects in the HCP-Aging dataset, with an age range of 36-89 years old (mean = 58.7, std = 14.3). The HCP-Young dataset was acquired with voxel size = 2 mm isotropic, TR = 720 ms, 1200 volumes, scan duration ≈ 14.4 min [29,30], and the HCP-Aging dataset was acquired with voxel size = 2 mm isotropic, TR = 800 ms, 487 volumes, scan duration ≈ 6.5 min [31,32].

### 2.2 Model Architecture

The two transformer-based architectures in our proposed framework are: seq2one (Figure 1a) and seq2seq (Figure 1b). Both of these architectures have two main components, a transformer block[25] followed by linear projection layers. The transformer block utilizes attention to extract temporal features from the input fMRI regions of interest (ROIs), which are passed to the linear projection layer(s) to predict RV and HR signals.

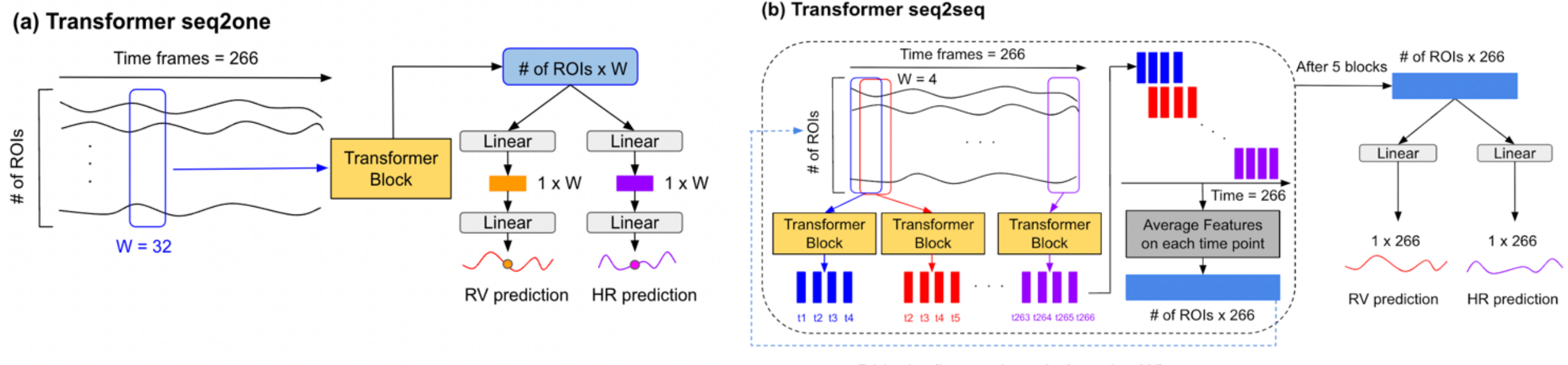

Figure 1. Transformer seq2one (a) and Transformer seq2seq (b) model architectures. These models take fMRI time-series from specific brain regions of interest (ROIs) as input, and output the predicted physiological (RV and HR) time courses.

For seq2one, we used a sliding-window approach (window size = 32 and step size = 1), similar to prior window-based CNN studies with younger-adult and pediatric cohorts [17,18]. For each fMRI window, we predict a single RV or HR time point in the middle of that window. For each time point, the transformer block (1 layer, 8 heads each with dimension 60, dropout rate = 0.3) learns to capture pairwise temporal dependencies between every time point within the same window. The extracted temporal features are then passed into linear projection layers to output a single RV or HR point. These individual points are assembled into the final predicted signals (235 time points), which are shorter than the input ROI signals (266 time points) due to edge effects at the first and last windows.

For seq2seq, we also adopted a sliding-window approach (window size = 4, 8, 12, 16, 20 and step size = ¼ of window size) [26–28]. For each fMRI window, the transformer block (20 heads each with dimension 100, dropout rate = 0.3) extracts temporal information and outputs features for each time point as the same length of the window. We then average the extracted features at each time point as a means of aggregating the overlapping features. As a result, each aggregated feature incorporates information from multiple windows that contain this time point, capturing more temporally global information. This aggregated feature is passed to the next transformer block, which has a larger window size. In this way, the model can leverage fine-grained features and longer-range temporal patterns. After 5 sequential seq2seq blocks, the final feature is passed to the linear projection layers, which output RV and HR signals that are the same length as the corresponding input ROI time series.

### 2.3 Training strategies and implementation details

Four training strategies were assessed: 1) directly applying the pre-trained model from HCP-Young to HCP-Aging, 2) training from scratch using only HCP-Aging data, 3) jointly training from scratch using both HCP-Aging and HCP-Young data and 4) pre-training with HCP-Young data and fine-tuning on HCP-Aging data. Models were implemented using PyTorch. The 5-fold train-test splits of the HCP-Aging dataset (described below) were done by adopting an age-balanced strategy. The validation set is selected as 15% of the training set scans. We chose Adam optimizer with default parameters with $\beta_1$=0.9, $\beta_2$=0.999, and batch size of 16. The model structures and hyperparameters for CNN and Bi-LSTM were directly adopted from previous papers [17,20]. For both Transformer models, the initial learning rate when training from scratch is 1e-4 with decay rate 0.5 and patience 2, and the initial learning rate for fine-tuning was 5e-5. Learning rate was chosen based on model performance on the HCP-Young dataset. Early-stopping was set to 5 on the validation set. Pearson correlation coefficient between the ground truth RV or HR and the predicted signals was used as the loss function and model evaluation metric. The experiments were performed on an NVIDIA GeForce RTX 2080 Ti.

### 2.4 fMRI and physiological signal preprocessing

Both fMRI datasets went through the HCP's generic fMRI volume minimal preprocessing pipeline, and were registered into MNI152 common space [30,33]. We extracted 400 cortical ROIs from the Schaefer Atlas [34], 25 subcortical ROIs from the Melbourne Atlas [35] and the Ascending Arousal Network Atlas [36] and 72 white matter ROIs from the Pandora Tractseg Atlas with a 0.95 threshold [37]. We further detrended the ROIs' time courses, bandpass filtered at 0.01 - 0.15 Hz, temporally resampled both datasets to 1.44 s and z-normalized each ROI's time course. The temporal downsampling step was carried out to more closely match fMRI data with conventional (typically > 1s) sampling interval (TR), and because the low-frequency oscillations in BOLD and RV/HR signals fall primarily within a frequency band that is adequately sampled by this rate. The physiological recordings of both the HCP-Young and HCP-Aging datasets were sampled at 400 Hz. RV was calculated as the standard deviation of a 6 s window centered at each TR, and HR was calculated in the same time window as the inverse of the mean inter-beat-interval. RV and HR were detrended, band-pass filtered (0.01 – 0.15 Hz) to extract low-frequency fluctuations, linearly resampled to 1.44 s, and z-normalized.

## 3. RESULTS

Table 1 compares the proposed framework with previously established models for RV and HR time course reconstruction on the HCP-Aging dataset. Values summarize the median Pearson's correlation across 5 folds. Predicted versus measured RV and HR time courses are presented for several older-adult subjects in Figure 2.

Table 1. RV and HR reconstruction performance on the HCP-Aging data. Values indicate Pearson's correlation between measured and recorded signals. TF: transformer

| **Task** | **RV Reconstruction** | | | | **HR Reconstruction** | | | |
|---|---|---|---|---|---|---|---|---|
| Model | CNN[17] | Bi-LSTM[20] | TF-seq2one | TF-seq2seq | CNN[17] | Bi-LSTM[20] | TF-seq2one | TF-seq2seq |
| # Parameters | 0.08 M | 40 M | 1.5 M | 22 M | 0.08 M | 40 M | 1.5 M | 22 M |
| Pretrain | / | 0.593 | / | / | / | 0.553 | / | / |
| From scratch | 0.619 | 0.665 | 0.685 | 0.670 | 0.506 | 0.592 | 0.585 | 0.602 |
| From scratch (jointly) | 0.592 | 0.657 | 0.684 | 0.675 | 0.530 | 0.593 | 0.593 | 0.605 |
| Fine-tune | 0.614 | 0.683 | **0.698** | 0.684 | 0.525 | 0.606 | 0.595 | **0.618** |

Overall, seq2one, when pretraining on the HCP-Young dataset and fine-tuning on the HCP-Aging dataset, achieved the best performance for RV reconstruction, and seq2seq achieved the best performance for HR reconstruction with the same training strategy. Notably, these models have fewer parameters than the Bi-LSTM model. Directly applying the pretrained model from HCP-Young data on the HCP-Aging data yielded the worst performance, indicating a domain shift between the young and old dataset. Training the model from scratch using HCP-Aging data yielded moderate performance, indicating that the model can effectively learn features from the aging group. Compared to training using only HCP-Aging data, jointly training from scratch with both HCP-Young and Aging data has comparable or decreased performance when reconstructing RV.

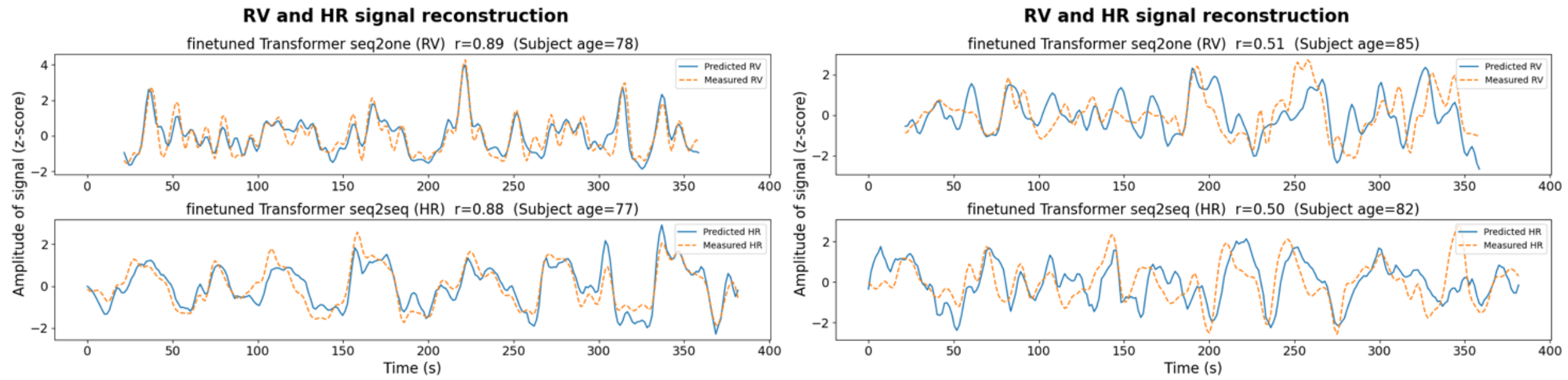

Figure 2. Examples of RV and HR signals predicted in older adults by the indicated models, together with the corresponding measured signals. Pearson correlation (r) between predicted and measured signals is also provided. The proposed frameworks effectively learn to predict many prominent features in these physiological time series.

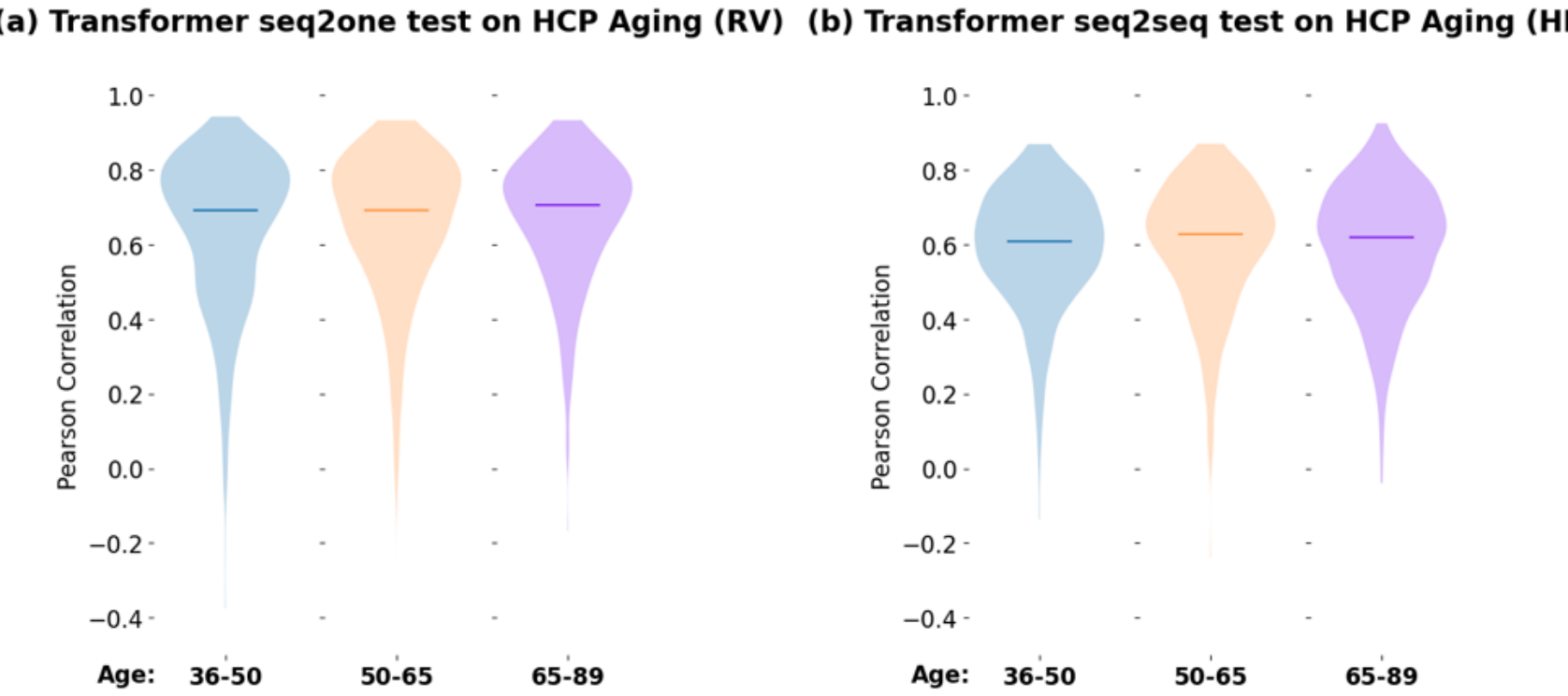

Figure 3. Performance across 3 age groups in the HCP-Aging data, for the indicated models.

When directly applying the Bi-LSTM model trained on HCP-Young dataset to the HCP-Aging dataset (without retraining), we observed a slight age-related decrease in model accuracy in older adults. Yet, when HCP-Aging data was included in the training set, no decreasing trend was found across different age groups (Fig. 3). This result may suggest that age-related information may be learned from the fMRI time courses without the explicit input of chronological age.

## 4. CONCLUSION

This work advances the ability to infer low-frequency cardiac and respiratory fluctuations directly from brain fMRI data. In a dataset from adults spanning a wide age range (36-89 years), the proposed transformer-based framework outperformed prior state-of-the-art on this task[17,19]. Further, the median performances on older adults (~0.6-0.7) are comparable to, or exceed, the performance on younger adults reported in prior studies [20]. Our experiments also revealed that leveraging additional data from young adults to pre-train the model, followed by fine-tuning with data from older adults, was helpful for capturing specific characteristics of the aging population. Overall, these methods allow for characterizing physiological components of fMRI data in older adults, and for investigating age-related changes in brain-body interaction and brain vascular health, in the common scenario of missing or corrupted physiological recordings.

**New/breakthrough work to be presented:** We proposed new frameworks with Transformer-based models to predict low-frequency cardiac and respiratory fluctuations from fMRI signals. Further, as the experiments in prior studies were limited to neuroimaging data from younger individuals (36 years and younger), this study presents a novel investigation into the older spectrum of the adult lifespan (ages 36-89), hence broadening the ability to model physiological features from fMRI.

## ACKNOWLEDGEMENTS

This work is supported by NIH RF1MH125931 and NIH P50MH109429. Research reported in this publication was supported by the National Institute on Aging of the National Institutes of Health under Award Number U01AG052564 and by funds provided by the McDonnell Center for Systems Neuroscience at Washington University in St. Louis. The HCP-Aging 2.0 Release data used in this report came from DOI: 10.15154/1520707. **This work has not been submitted elsewhere.**